# Automated Matchmaking to Improve Accuracy of Applicant Selection for University Education System

Oludayo, O. Olugbara
ICT & Society Research Group, Durban University of Technology, Durban, South Africa

Manish Joshi
Department of Computer Science, North Maharashtra University, Jalgaon MS, India

Michael M. Modiba
ICT & Society Research Group, Durban University of Technology, Durban, South Africa

Virendrakumar, C. Bhavsar
Faculty of Computer Science, University of New Brunswick, Fredericton, Canada

**Abstract**

The accurate applicant selection for university education is imperative to ensure fairness and optimal use of institutional resources. Although various approaches are operational in tertiary educational institutions for selecting applicants, a novel method – automated matchmaking – is explored in the current study. The method functions by matching a prospective student's skills profile to a programme's requisites profile. Empirical comparisons of the results, calculated by automated matchmaking and two other selection methods, show matchmaking to be a viable alternative for accurate selection of applicants. Matchmaking offers a unique advantage – it neither requires data from other applicants – nor compares applicants with each other. Instead, it emphasises norms that define admissibility to a programme. We have proposed the use of technology to minimize the gap between student's aspirations, skill sets and course requirements. It is a solution to minimize the number of students who get frustrated because of mismatched course selection.

## 1. Introduction

The selection of applicants for university education has shown to be a demanding obligation that does not consistently produce the expected results. The predicament of assigning suitable applicants to appropriate programmes puts a substantial weight on the shoulders of management of universities (Eisenkopf, 2009; Cardinal, Mousseau & Zheng, 2011). Over time, the exponential growth in the numbers of students that apply for admission at tertiary educational learning institutions as well as where issues of prior learning and matriculation results are taken into account exacerbated the problem (Mohtar, Zulkifli & Sheffield, 2011). Selection processes are generally only based on the institution's requirements – which expect from applicants to meet the admittance requisites. However, admission processes are reciprocal in nature. What is often overlooked is the *applicant's* perspective of the norms that universities apply when attracting prospective applicants. Hence, applicants may apply for admission at several universities, but will ultimately consider only the one that meets their personal and academic needs.

Factors influencing an applicant's decision for admission into universities include *inter alia* the reputation of the university, the cost of the study, its location, the applicant's need to study a particular programme, the family's preference for a particular field of study, peer pressure and the '*bandwagon*' effect. The integrity of the decision process – to admit an applicant – can be influenced by the selection criteria as well as the selection method implemented.

A myriad of measurement tools are available to tertiary educational institutions for designing criteria for the selection of first-time applicants, such as various standardised tests and in-depth interviews (Hardigan, Lai, Arneson & Robeson, 2001; Wilson, Chur-Hansen, Donnelly & Turnbull, 2008). However, inappropriate programme application requisites, poorly or obscurely stated entry requisites can regrettably contribute to decision-making errors. The need has arisen, therefore, to develop programme entry requisites more rigorous to avoid inaccuracies (Murray, Merriman & Adamson, 2008). An erratic applicant selection process is not only unethical, but also a waste of institutional resources (O'Neill, Korsholm, Wallstedt, Eika & Hartvigsen, 2009). In addition, an erroneous applicant selection process can understandably contribute to escalating skills mismatching, as an applicant may be selected for an inappropriate programme. Skills mismatch is understood as a breach that refers to knowledge, skills and competencies (Cedefop, 2009) and has recently received a lot of attention from various scholars, policy makers and governments (Desjardins & Rubenson, 2011; Mitrovic, Sharif, Taylor & Wesso, 2012). It has, therefore, become imperative to approach skills mismatching from the angle of striving for a much more accurate applicant selection.

Along this vein, the overarching objective of the current study was an attempt to improve the accuracy of the applicant selection process by means of a proposed matchmaking process that will provide an effective way to integrate cognitive and non-cognitive attributes to the applicant selection process. In this regard, Levin (2012) stresses the importance of non-cognitive skills in the assessment of educational outcomes. Accurate matchmaking can enable the selection process to be viewed from both the perspectives of applicants and institutions, since it allows for information of applicants and institutions to be incorporated in the selection process. In addition, matchmaking can benefit the central admission department of a university by enhancing the processing of applications – as well as assisting applicants to proactively determine programme choices – based on the recommendations generated by the proposed automated system.

## 2. Trends in Applicant-Selection Methods

Recently, research efforts have progressively been focussed on automating or simplifying the process of selecting suitable applicants for admission into higher educational institutions. In this section of the current research study, the authors report on selected scientific methods that had been implemented by other researchers – applying advanced technological systems – for the matching of a prospective applicant's skills profile and admission requirements into university's programmes. These methods can be classified into two main approaches: Statistical Machine Learning (SML) and Multi-criteria Decision Analysis (MDA).

Firstly, a system inspired by SML operates on a series of observed data samples by learning to perform a given task from the data samples. The SML methods are applied to resolve an applicant's selection problem as a binary classification task by predicting whether an applicant *qualifies* for a programme (forming a group of selected applicants) *or not* (a group of rejected applicants). The predictive value of SML methods lies therein that they contribute to improving the accuracy of an applicant selection process. However, the effectiveness of SML methods has been questioned because of the level of sophistication of the decision process, the assumptions made and the level of accuracy achieved (Lopez & Carlos, 2005). In addition, the methods utilise complex knowledge representation models, require more processing time, impose highly complex determinations to decision-makers and often do not clearly demonstrate the basis of ***how*** a qualifying applicant was selected (Mohtar, Zulkifli & Sheffield, 2011). SML methods – previously employed for applicant selection – include discriminate analyses (Graham, 1991); decision-tree methods (Ibrahim & Rusli, 2007); neural networks (Adewale, Adebiyi & Solanke, 2007); multivariate regression analyses (Huang & Fang, 2010); evolutionary algorithms (Shannon & McKinney, 2011) and fuzzy system analyses (Mohtar, Zulkifli & Sheffield, 2011).

Secondly, Multi-criteria Decision Analyses (MDA) are a class of multi-criteria optimization methods that make use of decision matrices to provide a systematic way for evaluating, or ranking, a set of alternatives (in this case applicants seeking admission into a university), relative to a set of decision criteria (the university's selection criteria). These decision criteria are usually associated with weights – as to reflect their relative importance. In the case of the application of MDA methods to practical decision problems, criteria weights are objectively determined by mathematical procedures, such as entropy (Deng, Yeh & Willis, 2000) or subjectively elicited from decision-makers. Furthermore, MDA methods follow the multi-dimensional characteristics of the applicant selection process in order to deal with the conflicting nature of multiple selection criteria. MDA methods are also suited to improve the applicant selection problem – as an evaluation task – by comparing applicants' academic performances with each other. In addition, the methods provide an effective framework for handling subjectivity and imprecision that may be present in the human decision-making process (Deng & Wibowo, 2009). However, the lack of domain dependency is a disadvantage of MDA methods. Moreover, criteria weights obtained by mathematical calculations might not accurately reflect the true preferences of individuals.

Thirdly, MDA methods include the following:
- the Electre III-genetic algorithm (Lopez & Carlos, 2005)
- Electre Tri models (Cardinal, Mousseau & Zheng, 2011)
- TOPSIS (Technique for Order Preference by Similarity to Ideal Solution) and SAW (Simple Attribute Weighting) (Manokaran, Subhashini, Senthilvel, Muruganandham & Ravichandran, 2011)
- Fuzzy-AHP-TOPSIS (Rana, Dey & Ghosh, 2012)

It is notable that both SML and MDA methods exclude the programme entry requisite dimensions to evaluate alternatives. This may result in an applicant – who did not pass a compulsory programme entry requisite – to be ranked higher, particularly when the applicant had not achieved high scores in other requisites. In addition, the two methods can only process numeric data. Non-numeric data have to be transformed into numeric data in order to qualify for processing. These restrictions to SML and MDA methods have necessitated the development of a simple, but the accurate matchmaking method in the educational domain.

At this point, it is necessary to discuss the Automated Matchmaking Method (AMM). The AMM fundamentally solves the applicant selection problem as a matchmaking task by calculating the similarity score between an applicant's skills profile and the profile of the programme requisites. The calculated similarity score yields the *requirement level* of an applicant for a given programme. Moreover, the AMM method does not require data from other applicants in order to evaluate an applicant. This feature makes the AMM method manageable. In addition, the method is able to directly process both numeric and non-numeric data.

## 3.    Automated Matchmaking Method

When applying the Automated Matchmaking Method (AMM), applicant selection is called *matchmaking*; a method that was originally employed in electronic marketplaces (Noia, 2004) and electronic learning (Liesbeth, Rosmalen, Sloep, Kon & Koper, 2007). Matchmaking performs as '*profiles of participants*' who are involved in the matchmaking process. Such a profile is a collection of a participant's expectations regarding products and services – offered or sought. For any profile '**P**', the matchmaking method locates the best available counterpart profile that matches the requirements specified in 'P'.

In the current study, institutions and applicants were the participants, having numerous and multifaceted expectations, called constraints. For matchmaking, various knowledge (rational) representation models were represented with the intention of exploring different types of constraints (Joshi, Bhavsar & Boley, 2009; Joshi, Bhavsar & Boley, 2010; Joshi, Bhavsar & Boley, 2011). In this regard, *knowledge* is a description of the '*real world*' and k*nowledge representation* is a way of encoding knowledge, in order to facilitate valid conclusions about the encoded knowledge.

The knowledge representation model – when applied to the matchmaking method – is formally represented as a set of constraints:

$$P = \{C_1, C_2, C_3 \dots C_m\} \qquad (1)$$

where each constraint is a quadruple of $Ci = <a, d, f, p>$; $a$ is an attribute; $d$ is a set of values for attribute value; $f$ is the flexibility that determines whether a constraint is *hard* or soft, and $p$ is the constraint priority.

In the context of university-applicant selection, *hard* constraints are programme entry requisites that *must* be fulfilled, whereas *soft* constraints are only desirable to be fulfilled. In a particular profile, flexibility $f$ assumes a Boolean value of 'Yes' or 'No', where 'No' corresponds to a *hard* constraint and 'Yes' means a *soft* constraint. The priority $p$ takes a value between '0' and '1' with '1' corresponding to the highest priority and '0' corresponding to the lowest.

## 4.    Composite Constraints

Textbox 1 shows a typical programme advertisement which represents a worldview of programme entry requisites, or a set of typical rules for the selection of applicants to a programme. This example of a worldview is used to illustrate the notion of composite constraints in programme-entry requisites.

*TEXTBOX 1: Programme entry requisites*

*Applications are invited for the full-time National Diploma in Information Technology, starting in January 2015. The university scholarship is available to full-time students. Computer programming skills or Web 2.0 technology skills will be an advantage. The*

> *programme can be accomplished as an on-campus or-off campus student. <u>A minimum of 50% in English, 50% in Mathematics and at least 40% in any three of Accounting, Afrikaans, Business Studies, Economics, Geography, Information Technology, IsiZulu, Physical Sciences and Venda is required.</u>*

The underlined segments in Textbox 1 correspond to 'composite constraints', which could not directly be modelled using a quadruple knowledge representation model. Therefore, the underlined requirements had to be enhanced using a scope operator denoted by '**::**' in order to accommodate the composite constraints. The scope operator was a resolution function that helped to identify and specify the category to which a constraint referred. A composite constraint was formally represented as <x::var, varDescrip, No, 1>, where 'x' was a category's name, 'x::var' was a keyword which elaborated that an attribute 'x::var' had 'varDescrip' as an attribute value. The 'var' could be a keyword 'count' indicating the minimum number of constraints to be satisfied among member constraints; 'x' and 'No' denoted the constraint flexibility and '1' the constraint priority.

*TEXTBOX 2: Applicant's skills*

> I seek admission to the full-time on-campus National Diploma in Information Technology programme starting in January. I have skills in computer programming and achieved the following school results: English 56%, Mathematics 65%, Geography 50%, IsiZulu 69%, Afrikaans 54%, Economics 62%, Physical Sciences 64% and Life Sciences 33%.

Textbox 2 shows the contents of an advertisement that had been exhibited. It contained a potential applicant's skills set – as a worldview of the knowledge of the applicant – seeking admission to a university.

*TEXTBOX 3: Knowledge representation of a programme requisites profile*

> <Compulsory_Subject::count,2,No,1>
> <Optional_Subject::count,>=3,No,1>
> <Compulsory_Subject::English_Language,50...100,No,1>
> <Compulsory_Subject::Mathematics,50...100,No,1>
> <Programme,National Diploma Information
> Technology,No,1>
> <Programme_Intake,January,No,1>
> <Programme_Type,Full-time,No,1>
> <Optional_Subject::Accounting,40...100,No,1>
> <Optional_Subject::Afrikaans,40...100,No,1>
> <Optional_Subject::Business_Studies,40...100,No,1>
> <Optional_Subject::Economics,40...100,No,1>
> <Optional_Subject::Geographic,40...100,No,1>
> <Optional_Subject::Information_Technology,40...100,No,1>
> <Optional_Subject::IziZulu,40...100,No,1>
> <Optional_Subject::Physical_Sciences,40...100,No,1>
> <Optional_Subject::Venda,40...100,No,1>
> <Residence,{On Campus, Off Campus},No,1>
> <Scholarship, Yes, No,1>
> <Skills, {Web 2.0 Technology, Computer
> Programming},No,1>

Textbox 3 displays the knowledge representation model of the programme entry requisites in Textbox 1. The <Compulsory_Subject**::**count,2,No,1> and <Optional_Subject**::**count,>=3,No,1> composite constraints were added to the profile, indicating that the *two* constraints in the 'Compulsory_Subject' category and any *three* constraints in the 'Optional_Subject' category, were the minimum programme entry requisites.

*TEXTBOX 4: Knowledge representation of an applicant's skills profile*

<Programme, National Diploma Information
Technology,No,1>
<Programme_Intake,January,No,1>
<Programme_Type, full-time, No, 1>
<Residence, On Campus,No,1>
<Scholarship, Yes,No,1>
<Afrikaans,54,No,1>
<English_Language,56,No,1>
<Geographic,50,No,1>
<Economics,62,No,1>
<IsiZulu,69,No,1>
<Life_Sciences,33,No,1>
<Mathematics,65,No,1>
<Physical_Sciences,64,No,1>
<Skills, Computer Programming,No,1>

Textbox 4 shows by what means the applicant's skills profile – shown in Textbox 2 – was represented, using the knowledge representation model of matchmaking. The composite constraint count – at the level of the programme requisites profile – was explicitly specified, but automatically determined for an applicant's skills profile. Moreover, it was not mandatory for the user of the automated matchmaking system to specify the composite constraints in an applicant's skills profile. This eliminated some skilful efforts for the system users who might have been interested in querying the programme profile dataset, but were unfamiliar with the identification of the conventions in the different profiles.

## 5.    Profiles Preprocessing

The profiles preprocessing algorithm constructed two non-composite constraint profiles 'P*c' and 'P*a' from the programme requisites profile ('Pc') and the applicant's skills profile ('Pa') respectively. These constructed profiles served as input to the matchmaking algorithm (Joshi, Bhavsar & Boley, 2010) with the aim of calculating their *similarity score*. Moreover, the profiles preprocessing algorithm produced the *similarity score* between the two input profiles. The objective of this procedure was to preserve the original contents of the participant profiles. In this regard, the underlying principle of profiles preprocessing was to replace the composite member constraints of a category – with a single non-composite constraint – called a target constraint. Next, the attribute values of the target constraints were determined as to accomplish this replacement. For each category in 'Pc', two target constraints were constructed to bear the category name, but they could possibly have different attribute values. One target constraint was inserted into 'P*c' and the other into 'P*a'. It is important to note that the way the attribute value of a target constraint is determined, can influence the matchmaking result. The composite constraint count of an applicant skills profile was the number of constraints in the skills profile that matched the constraints in a programme requisites profile (Joshi, Olugbara, Bhavsar, Lall & Modiba, 2012).

The current research practice uses the sum of attribute values of composite constraints to represent the attribute value of a target constraint. This improves the discriminating power of the matchmaking algorithm. Accordingly, if two applicant skills profiles have the same composite constraint count, their similarities to a programme requisites profile will differ, provided they have differing attribute values. Consequently, an accurate list of ranked programme recommendations will be generated – using the matchmaking algorithm.

The attribute values of target constraints were scaled to eliminate the effects of varying the number of member constraints. The current authors supposed that x = (x1, x2, …, xM) was the set of attribute values of M composite constraints in a category 'c' in a profile 'Pc' while y = (y1, y2, …, yN) was the set of attribute values of N constraints in a profile 'Pa' – in such a way that the constraints were in a category 'c' of a profile 'Pc'. Accordingly, all composite constraints of category 'c' in 'Pc' were replaced by a target constraint with attribute 'c' – of which the attribute value was ' $Pcc$ ' and determined as follows:

$$Pcc = \left( \frac{S}{M} \right) \sum_{i=1}^{M} xi$$

(2)

The parameter S was the specified count of constraints to be fulfilled among M ≥ S member constraints. In Textbox 3 for example, by considering the category Optional_Subject, the value of the parameters S and M are S = Optional_Subject::count = 3 and M = 9. The corresponding target constraint – to replace all constraints in a profile 'Pa' that were found in the category 'c' of a profile 'Pc' – would bear the attribute name 'c'. The attribute value ' $Pac$ ' of the target constraint was determined as follows:

$$Pac = \left( \frac{\min(S,T)}{\max(S,N)} \right) \sum_{j=1}^{N} yi$$

(3)

The min(x,y) and max(x,y) functions, respectively, returned the minimum and the maximum values of their arguments x and y. The parameter T was the count of constraints in 'Pa' that were satisfied in a category 'c' of 'Pc' – and S ≤ N ≤ M was the count of constraints in 'Pa' that were in the category 'c' of 'Pc'. In Textbox 4, T = 5 and N = 5. The attribute 'Life_Sciences' did not contribute to the determination of Pac because it was not in the Optional_Subject category.

It is imperative, in this regard, to note that the attribute values were calculated for numeric and range types, but determined for non-numeric types as $Pcc = S / M$ and $Pac = \min(S,T) / \max(S,N)$. The profiles preprocessing algorithm is shown in Textbox 5, wherein the function insert (Ci, Pi) inserted the constraint 'Ci' into a profile 'Pi'.

**TEXTBOX 5 Profiles preprocessing algorithm**

```
Algorithm profiles Preprocessing()
Input parameters: Pj, Pi
Auxiliary variables: Pj*, Pi*
Output variable: Similarity
1.Foreach category 'c' in 'Pc' do
    1.1 Find Pcc by using Equation (2) or  Pcc = S / M
    1.2 Extract the constraints of the category 'c' in 'Pa'
    1.3  Find   Pac   by   using   Equation   (3)   or
Pac = min(S,T) / max(S,N)
    1.4 insert(<c, Pcc, No, 1>, Pc*)
    1.5 insert(<c, Pac, No, 1>, Pa*)
2. For each non-composite constraint Cc in 'Pc', insert(Cc,Pc*)
3. For each non-extracted constraint Ca in 'Pa', insert(Ca,Pa*)
4. Find Similarity=matchmaking(P*a, P*c)
5. Output Similarity
End_Algorithm
```

## 6.      Matchmaking Algorithm

The matchmaking algorithm (Joshi, Bhavsar & Boley, 2010) compares two constraints of two profiles to obtain a match. If the constraint attributes are *equal*, an intermediate similarity value is calculated by checking the attribute values.  If the attribute values are *different*, an intermediate similarity value is calculated by considering the constraint flexibility values. The similarity between two attribute values is compared whenever 'hard' constraints are mismatched. In the present study the similarity $S(Pi,Pj)$ between the profiles $Pi$ and $Pj$ was calculated – using the similarity $S_k(Ci,Cj)$ between the constraints $Ci$ and $Cj$ as:

$$S(Pi,Pj) = \prod_{k=1}^{N} S_k(Ci,Cj) \qquad (4)$$

Generally, $S_k(Ci,Cj)$ can be defined for vectors X and Y using $\lambda,\mu$–normalization $[0, 1/\mu]$ bounded distance function (Yianilos, 1991) as:

$$S_k(X,Y) = 1 - \frac{\| X - Y \|}{\lambda + \mu(\| X \| + \| Y \|)} \qquad (5)$$

In the present study, $\lambda = \mu = 1$ and the norm function could be taken as a city block or Euclidean distance function with the aim of calculating the similarity between vectors 'X' and 'Y'.

## 7.      Results and Discussion

In the study at hand, the automated matchmaking method (AMM) was applied to real data obtained from applicants at the Durban University of Technology (DUT) in South Africa. The DUT advertises their programmes (Textbox 1) through radio, visits to schools and open weeks – where the university showcases their various programmes to invited schools. Applicants  are required to apply via the central admission office (CAO) for a place at DUT. The selection officer (academic staff) in the department would then be able to access all the CAO applications of the applicants who had applied for a particular programme. The selection officer could also instantly see whether an applicant had met the basic entry requirements for the programme. The selection officer could rank the applicants from highest to lowest - according to the total scores obtained (Rank) and whether they qualified for selection (Selected) as in Table 1. The regulation at the DUT is that a secure proposition can only be offered to applicants who have already completed their matriculation examination. Applicants are then sent letters from the institution informing them whether they have been accepted, or not.

TABLE 1: Performance data for initiating the applicant selection process (%)

| Applicant | Sub1 | Sub2 | Sub3 | Sub4 | Sub5 | Sub6 | Total | Rank | Selected |
|-----------|------|------|------|------|------|------|-------|------|----------|
| 001 | 47 | 72 | 68 | 84 | 75 | 87 | 433 | 23 | No |
| 002 | 48 | 80 | 66 | 71 | 85 | 64 | 414 | 22 | No |
| 003 | 49 | 87 | 70 | 65 | 86 | 46 | 403 | 21 | No |
| 004 | 68 | 67 | 62 | 75 | 69 | 49 | 390 | 1 | Yes |
| 005 | 59 | 66 | 68 | 73 | 69 | 53 | 388 | 2 | Yes |
| 006 | 59 | 66 | 55 | 89 | 73 | 46 | 388 | 2 | Yes |
| 007 | 45 | 64 | 64 | 70 | 76 | 67 | 386 | 24 | No |
| 008 | 52 | 69 | 65 | 67 | 68 | 61 | 382 | 4 | Yes |
| 009 | 68 | 58 | 52 | 71 | 63 | 68 | 380 | 5 | Yes |
| 010 | 70 | 71 | 58 | 67 | 67 | 46 | 379 | 6 | Yes |
| 011 | 53 | 71 | 64 | 69 | 74 | 47 | 378 | 7 | Yes |
| 012 | 67 | 80 | 65 | 51 | 58 | 57 | 378 | 7 | Yes |
| 013 | 52 | 69 | 66 | 71 | 73 | 46 | 377 | 9 | Yes |
| 014 | 53 | 64 | 46 | 78 | 56 | 77 | 374 | 10 | Yes |
| 015 | 57 | 62 | 49 | 80 | 75 | 50 | 373 | 11 | Yes |
| 016 | 60 | 63 | 52 | 93 | 59 | 46 | 373 | 11 | Yes |
| 017 | 57 | 74 | 59 | 74 | 65 | 44 | 373 | 11 | Yes |
| 018 | 67 | 69 | 45 | 66 | 70 | 55 | 372 | 14 | Yes |
| 019 | 55 | 78 | 59 | 70 | 72 | 38 | 372 | 14 | Yes |
| 020 | 67 | 62 | 60 | 67 | 60 | 55 | 371 | 16 | Yes |
| 021 | 62 | 76 | 65 | 42 | 42 | 77 | 364 | 17 | Yes |
| 022 | 55 | 74 | 53 | 71 | 65 | 44 | 363 | 18 | Yes |
| 023 | 69 | 60 | 56 | 66 | 51 | 60 | 362 | 19 | Yes |
| 024 | 50 | 75 | 50 | 74 | 63 | 49 | 362 | 19 | Yes |
| 025 | 45 | 71 | 60 | 67 | 75 | 40 | 359 | 25 | No |

Table 1 shows the results of the 25 applicants randomly selected from the list of 5000 applicants who had been appraised on the basis of their matriculation results. The applicants needed to pass the Mathematics (Sub1) and English language (Sub2) as two compulsory subjects - in addition to any three subjects among Sub3 to Sub6 – in order to qualify for selection. It can be seen from Table 1 that 20 applicants – who had met the programme entry requisites – were selected and five applicants were rejected, as they had failed to obtain a minimum score of 50% in Sub1.

TABLE 2: Comparative results of the evaluation scores and ranks calculated by SAW, the TOPSIS and the AMM

| Applicant | SAW | TOPSIS | AMM |
|-----------|-----|--------|-----|

| | | | | | | |
|---|---|---|---|---|---|---|
| 001 | 0.86182 | 1 | 0.66258 | 1 | 0.40527 | 23 |
| 002 | 0.83560 | 2 | 0.58544 | 2 | 0.40605 | 22 |
| 003 | 0.82415 | 3 | 0.51397 | 4 | 0.40794 | 21 |
| 004 | 0.80697 | 4 | 0.50852 | 5 | 0.62836 | 1 |
| 005 | 0.79548 | 6 | 0.49331 | 9 | 0.61116 | 4 |
| 006 | 0.78834 | 8 | 0.49338 | 8 | 0.61116 | 4 |
| 007 | 0.77382 | 11 | 0.49980 | 7 | 0.35246 | 24 |
| 008 | 0.77832 | 10 | 0.48230 | 11 | 0.59628 | 9 |
| 009 | 0.78069 | 9 | 0.51915 | 3 | 0.60133 | 7 |
| 010 | 0.79174 | 7 | 0.47520 | 14 | 0.61781 | 3 |
| 011 | 0.77318 | 12 | 0.44560 | 19 | 0.59546 | 10 |
| 012 | 0.79632 | 5 | 0.48423 | 10 | 0.62114 | 2 |
| 013 | 0.76977 | 15 | 0.43898 | 20 | 0.58934 | 13 |
| 014 | 0.75337 | 22 | 0.50752 | 6 | 0.57848 | 16 |
| 015 | 0.75644 | 20 | 0.45481 | 17 | 0.58048 | 15 |
| 016 | 0.75990 | 19 | 0.45629 | 16 | 0.58678 | 14 |
| 017 | 0.76799 | 16 | 0.43388 | 21 | 0.59762 | 8 |
| 018 | 0.76985 | 14 | 0.47535 | 13 | 0.60164 | 6 |
| 019 | 0.76688 | 17 | 0.42599 | 22 | 0.45626 | 20 |
| 020 | 0.77007 | 13 | 0.45710 | 15 | 0.59209 | 11 |
| 021 | 0.76490 | 18 | 0.47683 | 12 | 0.59047 | 12 |
| 022 | 0.74458 | 23 | 0.39796 | 24 | 0.57822 | 17 |
| 023 | 0.75393 | 21 | 0.44709 | 18 | 0.57822 | 17 |
| 024 | 0.73614 | 24 | 0.40134 | 23 | 0.57182 | 19 |
| 025 | 0.72821 | 25 | 0.37760 | 25 | 0.34079 | 25 |

Table 2 discloses the results of the calculations by means of the AMM – compared to the results of the Simple Attribute Weighting (SAW) (Afshari, Mojahed & Yusuff, 2010) – as well as the Technique for Order Preference by Similarity to Ideal Solution (TOPSIS) (Hwang, Lai & Liu, 1993). Both the SAW and the TOPSIS methods are shared examples of MDA methods. The set of relative weights of the selection criteria used in the SAW and the TOPSIS calculations was obtained from the selection officer: {0.20; 0.20; 0.15; 0.15; 0.15, 0.15}. The weight function implied that Sub1 and Sub2 had equal weights of 20% and each of the remaining subjects (Sub3 to Sub6) were weighted at 15%. In the application of the AMM, a programme requisites profile (Textbox 3) was built using the information from the programme eligibility norms (Textbox 1).

In addition, a skills profile (Textbox 4) was built for each applicant using the data in Textbox 2. The evaluation score, computed by the AMM, was considerably reduced when an applicant failed to satisfy a mandatory subject (Sub1 or Sub2). For instance, among the five applicants 001, 002, 003, 007 and 025 that were rejected, applicants 001, 002, 003 and 007 were ranked highly by SAW and the TOPSIS. However, the AMM ranked these four applicants lowly for not satisfying 'Sub1'. Applicant 004 obtained the highest AMM evaluation score (0.62836) and was subsequently ranked no.1 by the AMM. Similarly, applicant 025 with the lowest ranking score (0.34079), was the worst-performing applicant and was ranked 25[th] by all three methods. The result (Table 2) also shows that when the AMM was applied for selection, the applicant who had received the highest ranking score was also the most qualified for the selection.

FIGURE 1: System rankings against human rankings

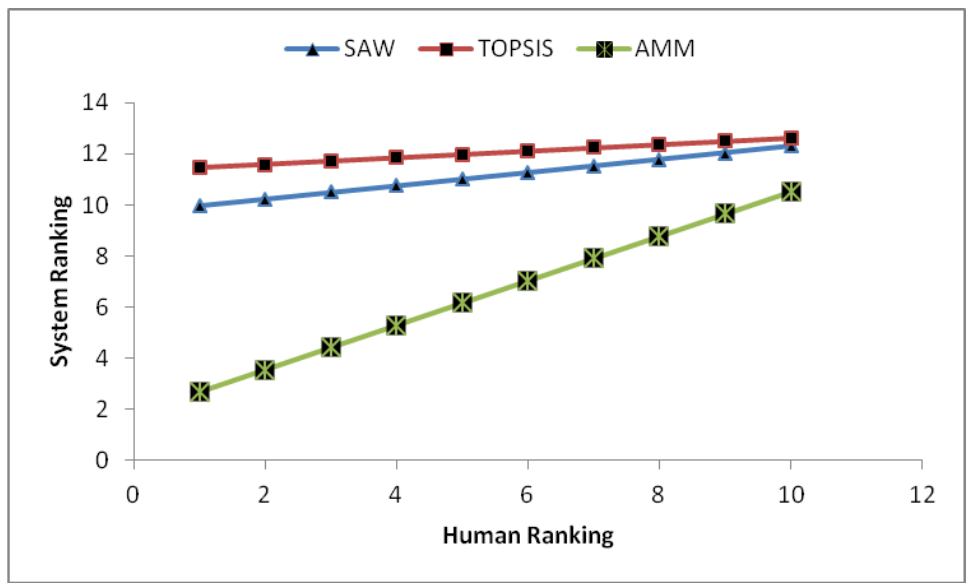

Figure 1 shows the regression graph with three rankings that were calculated by the three automated systems (SAW, TOPSIS and AMM) against the ranking that was assigned to an applicant by the selection officer (human ranking). The human ranking is determined by the total score obtained from six subjects in the matriculation examination (Table 1). According to the results in Figure 1, the AMM was the most accurate among the three evaluation methods and it confirms the evaluation, ranking and selection of most of the qualified applicants. The worldview of applicant selection – as generated by the AMM – is seen to be close to the worldview as generated by the human ranking. The Spearman-product correlation coefficients of the regression rankings were 0.133 (F-statistic = 0.413); 0,259 (F-statistic = 1.660) and 0.878 (F-statistic = 77.07) for the TOPSIS, SAW and the AMM, respectively.

Additionally, the F-statistic value of 77.607 for the AMM was much higher than the critical F-value of 3.420. Consequently, on a 95% level of confidence, the ranking generated by the AMM was not a random scatter point – therefore the regression rankings were justified. The big improvement in the calculation of the evaluation scores by the AMM was the result – or the method's ability – to match an applicant's skills profile against a programme requisite's profile. Moreover, this result indicates that the accuracy of the applicant selection process was effected by the selection method implemented.

## 8.    Conclusion

Admission selection processes naturally involve a number of important considerations, such as the determination of valid programme entry requisites and an appropriate selection method. The results from the study at hand demonstrate the usefulness of automated matchmaking as a method of improving the accuracy of an applicant's selection process for admission to a university in South Africa. The inference can be made that the AMM offers a university a useful tool to automate the process of evaluation of students and – if required – to improve an applicant's selection. The method generates a worldview of an applicant's selection process that is close to the human view. The method is relatively unassuming and flexible with the aim of accurately select the best-qualified applicants. In future, if cognitive and interpersonal abilities of students can be measured, it can be easily modelled in the system. Moreover, it is the strong point of the system. If a student completed some certificate course in say 'Inter-person relationship' it can be easily added to the skills set of the student and can be represented in the form of quadruple.

In addition, the AMM provides insight into the *accuracy* of an applicant's selection process, which can naturally be influenced by the selection method itself. However, the result – calculated by the AMM – could be affected by the *suitability* of the selection criteria that are being applied. Admission standards continue to be the sole responsibility of a university in order to implement effective programme entry requisites – otherwise fairness and quality are compromised in the selection process.

The unique contribution of the current research project lies in the exploration of *automated matchmaking,* which proved to be a viable alternative to the more 'conventional' admission processes. In addition, the implementation of *automated matchmaking* is directly related to applicant selection. It can also be used in other application protocols, as well as other types of selection approaches. For instance, it can, among others, be applied to staff recruitment, library resource procurement, learning object selection, assignment of subjects to lecturers, the selection of the best performing students for an award, as well as the assigning of research students to supervisors. These are some of the applications where decision requirements can be represented as constraints, following the knowledge representation profile of the matchmaking method.

In a future research project, the AMM will be extended to aid applicant selection in the presence of imprecise data. In addition, it is prudent to explore the role of domain ontology in the matchmaking process.